\documentclass[runningheads]{llncs}

\usepackage{multirow}
\usepackage{colortbl}
\usepackage{amssymb}%
\usepackage{pifont}%
\usepackage{eccv}

\usepackage{eccvabbrv}

\usepackage{graphicx}
\usepackage{booktabs}
\usepackage{array}
\usepackage{wrapfig}
\usepackage{algorithmic}
\usepackage[ruled, linesnumbered]{algorithm2e}
\usepackage{etoolbox}
\usepackage[accsupp]{axessibility}  %
\newcommand{\xmark}{\ding{55}}
\newcommand{\cmark}{\ding{51}}
\newcommand{\bestresult}[1]{\textbf{\textcolor{red}{#1}}}
\newcommand{\secondbest}[1]{\textcolor{blue}{\underline{#1}}}
\newcommand{\supp}[1]{{\color{blue}  #1}}

\newcommand{\venueTT}[1]{{$_{\texttt{\text{#1}}}$}}

\usepackage{hyperref}

\usepackage{orcidlink}

\begin{document}

\title{FinePseudo: Improving Pseudo-Labelling through Temporal-Alignablity for Semi-Supervised Fine-Grained Action Recognition}

\titlerunning{FinePseudo: Semi-Supervised Fine-Grained Action Recognition}

\author{Ishan Rajendrakumar Dave\orcidlink{0000-0001-9920-6970} \and Mamshad Nayeem Rizve\orcidlink{0000-0001-5378-1697} \and Mubarak Shah\orcidlink{0000-0001-6172-5572}}

\authorrunning{Dave et al.}

\institute{Center for Research in Computer Vision,
University of Central Florida, USA\\
\email{ishandave@ucf.edu}, \email{nayeemrizve@gmail.com}, \email{shah@crcv.ucf.edu}\\
\url{https://daveishan.github.io/finepsuedo-webpage/}
}

\maketitle

\begin{abstract}
  Real-life applications of action recognition often require a fine-grained understanding of subtle movements, e.g., in sports analytics, user interactions in AR/VR, and surgical videos. Although fine-grained actions are more costly to annotate, existing semi-supervised action recognition has mainly focused on coarse-grained action recognition. Since fine-grained actions are more challenging due to the absence of scene bias, classifying these actions requires an understanding of action-phases. Hence, existing coarse-grained semi-supervised methods do not work effectively. In this work, we for the first time thoroughly investigate semi-supervised fine-grained action recognition (FGAR). We observe that alignment distances like dynamic time warping (DTW) provide a suitable action-phase-aware measure for comparing fine-grained actions, a concept previously unexploited in FGAR. However, since regular DTW distance is pairwise and assumes strict alignment between pairs, it is not directly suitable
  for classifying fine-grained actions. To utilize such alignment distances in a limited-label setting, we propose an Alignability-Verification-based Metric learning technique to effectively discriminate between fine-grained action pairs. Our learnable alignability score
  provides a better phase-aware measure, which we use to refine the pseudo-labels of the primary video encoder. Our collaborative pseudo-labeling-based framework `\textit{FinePseudo}' significantly outperforms prior methods on four fine-grained action recognition datasets: Diving48, FineGym99, FineGym288, and FineDiving, and shows improvement on existing coarse-grained datasets: Kinetics400 and Something-SomethingV2. We also demonstrate the robustness of our collaborative pseudo-labeling in handling novel unlabeled classes in open-world semi-supervised setups.

  \keywords{Fine-grained Action Recognition \and Semi-supervised learning}
\end{abstract}

\section{Introduction}
Considering the action recognition problem in practice, many critical applications demand high precision in classifying subtle movements. For instance, in analyzing surgical videos to monitor subtle patient movements~\cite{tscholl2020situation}, AR and VR applications~\cite{xu2021hmd}, require identifying the user's nuanced movements for a more responsive interaction, and in sports analytics~\cite{hong2021video,naik2022comprehensive}, it enables detailed action quality assessment and injury prevention.

\begin{figure}[t]

    \centering
    \includegraphics[width=\textwidth]{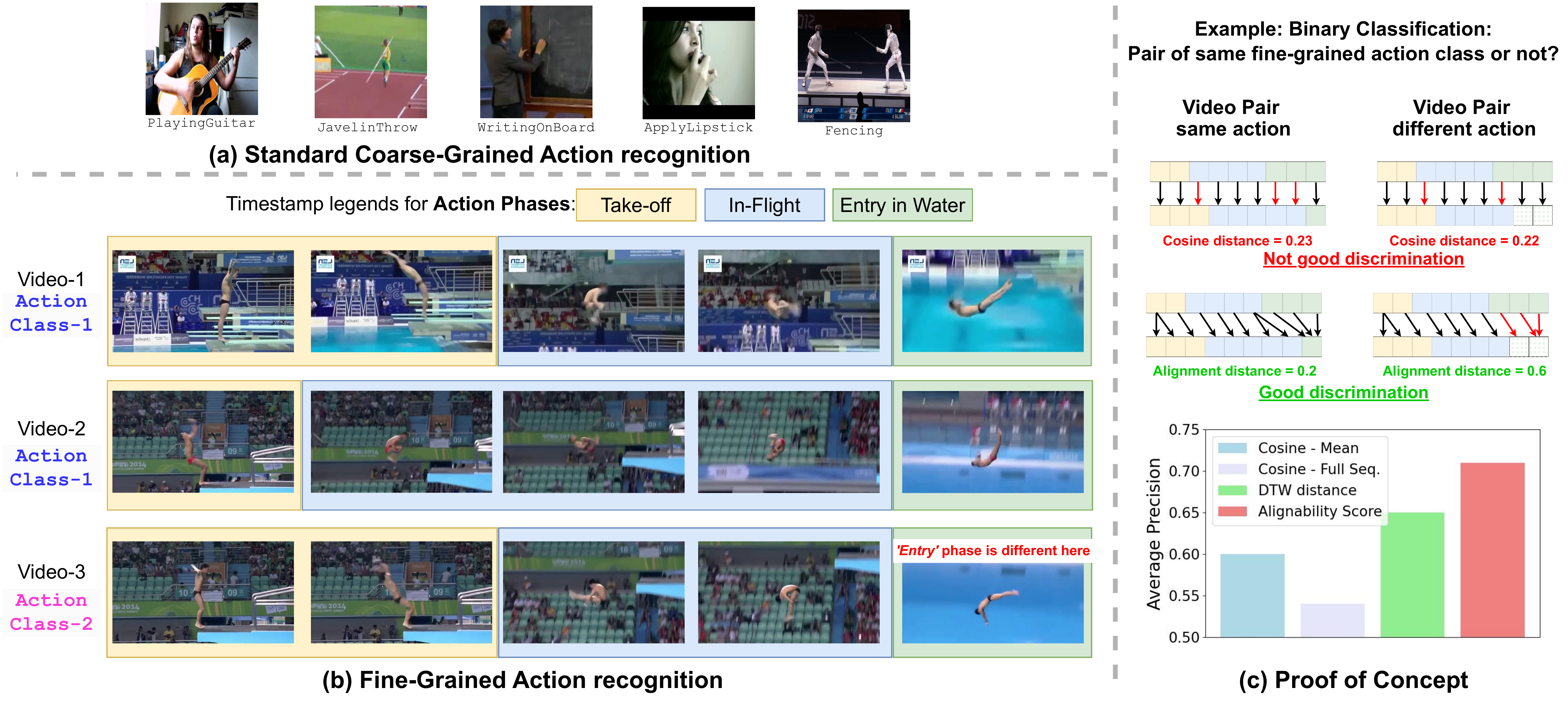}
    \caption{
\textbf{(a)} Sample actions from standard coarse-grained action recognition dataset (UCF101) \textbf{(b)} Sample actions from fine-grained action recognition dataset (Diving48) \textbf{(c)} For proof-of-concept, we choose a binary classification problem of fine-grained actions, where the model has to predict whether the pair of videos belong to the same class or not. We consider Diving48 dataset with 10\% training data. We first obtain the frame-wise video embedding from a pretrained framewise video-encoder model (Details in Sec.~\ref{sec:algo}). The top part of (c) shows that the cosine distance computed at each timestamp does not provide a discriminative measure, whereas, DTW-based alignment cost provides a clear difference in pair of same vs different classes. The bottom part of (c), shows the performance of the binary classification task in terms of average precision, where our alignability-score significantly outperforms the other standard distances.}
    \label{fig:teaser_a}
    \vspace{-3mm}
\end{figure}

Although fine-grained action recognition (FGAR) allows for wider adoption of %
action recognition in real-life applications, research has mainly focused on coarse-grained action recognition~\cite{hvu,hmdb,hacs,activitynet,charades}. For instance, from Fig.~\ref{fig:teaser_a}(a), we observe that coarse-grained action covers broader classes, such as `PlayingGuitar' vs `JavelinThrow'. Subtle human movements are not essential for classifying these, given their very different motion pattern and inherent scene bias (i.e., the scene provides substantial cues for identifying action)~\cite{choi2019can}. Conversely, Fig.~\ref{fig:teaser_a}(b) illustrates fine-grained action categories from `Diving', comprising \textit{action-phases} like `Take-off', `In-flight', and `Entry into Water'. This figure demonstrates that even a difference in the `Entry' phase from video-2 to video-3 alters the action class from class-1 to class-2. 
This suggests that FGAR can significantly benefit from learning action-phases. %

However, annotating such fine-grained actions poses significant challenges. Unlike coarse-grained actions, fine-grained actions require extensive, often repetitive viewing and expert annotation, making the process time-consuming and costly. This underscores the need for a semi-supervised learning approach for FGAR. However, current semi-supervised methods, designed for broader action categories, heavily rely on complex augmentation schemes like strongly and weakly augmented versions~\cite{semi_cmpl}, token-mix~\cite{xing2023svformer}, or actor-cutmix augmentations~\cite{semi_actorcutmix}. These techniques, while successful in standard datasets mainly for %
exploiting scene bias, may not be effective for FGAR due to scene uniformity across actions. Moreover, recent video-level self-supervised methods~\cite{timebalance}, successful in limited data contexts, do not effectively capture frame-level changes in action phases, which is crucial for recognizing fine-grained actions.

To build a solution tailored for fine-grained action recognition, we conduct a preliminary study to better understand the efficacy of different distance metrics in differentiating fine-grained videos. Let's take the example of binary classification of fine-grained actions, shown in Fig.~\ref{fig:teaser_a}(c). Here, the goal is to verify whether the two videos belong to the same or different class utilizing the embeddings from a frame-wise video encoder ($f_A$) in a limited labeled data setting. Our experiments demonstrate that standard feature distances like cosine distance are inadequate for this classification task. 
Particularly, we notice that computing cosine-distance over the temporally pooled features loses the temporal-granularity whereas computing cosine distance on the temporally unpooled features is %
suboptimal since different action phases take different amounts of time%
, e.g., phases in video-1 and video-2 of Fig.~\ref{fig:teaser_a}(b).
Therefore, a better way to compute the distance between a pair of fine-grained actions should be done by making \textit{phase-to-phase} comparisons. One way to obtain such phase-aware distance is by aligning the phases of the video embeddings. 
Hence, we %
hypothesize that alignability (i.e., whether two videos are alignable or not) based verification can provide a better phase-aware solution to differentiate fine-grained actions. %

One way to achieve such phase-aware distance is through alignment distance - dynamic time warping (DTW) optimal path distance. We see a significant boost in class-discrimination capability over regular cosine distances as shown in Fig.\ref{fig:teaser_a}(c) bar chart. This observation has not been explored before to solve FGAR in the limited labeled setting. At the same time, standard DTW distance is not an ideal %
class-discriminative measure as its optimal path assumes strict alignment between two videos and the final distance depends on the length of the video and frame-level similarities. Based on this observation, we propose an \textit{alignability-verification}-based metric learning technique to learn from the labeled data and produce a \textit{learnable alignability score} for a pair of videos. In the chart Fig.\ref{fig:teaser_a}(c), we see that our learnable alignability score improves the class-discriminative capability of DTW and provides a better distance measure for discriminating a pair of fine-grained videos.

Once such limited-labeled training is completed, we can utilize this alignability metric for pseudo-labeling (PL) by producing class-wise alignability-scores. These temporal aliganbility based pseudo-labels provide complementary information to the standard pseudo-labels generated from output confidence scores. To benefit from these complementary sets of pseudo-labels, we employ a collaborative pseudo-labeling process for semi-supervised fine-grained action recognition. Particularly, we combine the class predictions from frame-wise encoder, $f_A$, and finetuned video encoder, $f_E$, to get a refined pseudo-label. We update these pseudo-labels iteratively and conduct training in a self-training framework.

\noindent{The major contributions of this work are summarized as follows:}

\begin{itemize}

\item Our work is the first %
to thoroughly study the problem of fine-grained semi-supervised action recognition. We present \textit{FinePseudo}, a co-training framework where we utilize temporal-alignability to improve the pseudo-labeling process of unlabeled fine-grained videos.

\item To learn effectively from the limited labeled fine-grained videos, we propose a alignability-verification-based metric learning technique.
\item For collaborative pseudo-labeling, we design a non-parametric classifier-based prediction from the learnable alignability scores to refine output predictions. %

\item We evaluate our method on 4 fine-grained action recognition datasets: Diving48, FineGym99, FineGym288, and FineDiving, where our method significantly outperforms prior semi-supervised action recognition methods. Our method also performs competitively %
against the prior methods on coarse-grained datasets like Kinetics400 and Something-SomethingV2.

\item We demonstrate the robustness of our collaborative pseudo-labelling method in handling novel unlabeled classes in open-world semi-supervised setups.

\end{itemize}

\section{Prior Work}
\noindent\textbf{Semi-supervised Action Recognition}
Semi-supervised learning is still a growing area of research for action recognition compared to the image domain~\cite{zheng2022simmatch,cai2022semi,wang2022np,arazo2020pseudo,zhang2021flexmatch,pham2021meta,simclrv2,yang2022class,assran2021semi,rizve2022openldn,rizve2022towards}. To exploit the additional temporal dimension, various methods have employed additional modalities, including temporal gradients~\cite{semi_tgfixmatch}, optical-flow~\cite{semi_mvpl}, and P-frames~\cite{semi_compressed}. Concurrently, interesting augmentation schemes have been proposed, such as slow-fast streams~\cite{semi_tacl}, strong-weak augmentations~\cite{semi_cmpl}, CutMix~\cite{semi_actorcutmix}, and token-mix~\cite{xing2023svformer}. While ~\cite{timebalance} shows the potential of self-supervised video representations (videoSSL) in leveraging the unlabeled videos for semi-supervised setup.

However, these approaches mainly address semi-supervised action recognition problems focusing on coarse-grained actions with significant scene bias~\cite{choi2019can}, where the scene context provides substantial cues for action recognition. For fine-grained actions, which typically occur within the same scene, methods tailored for scene-bias datasets may not be fully applicable. For instance, augmentations like token-mix or CutMix might lose their effectiveness in uniform scene environments. Similarly, some methods may be partially ineffective, such as the appearance branch of~\cite{semi_mvpl}, or the temporally-invariant teacher of~\cite{timebalance}. While approaches like~\cite{semi_tcl} have shown results on ~\cite{goyal2017something}, their application has not been thoroughly explored beyond human-object interaction, leaving a gap in addressing diverse human actions.

Motivated by these gaps, we propose, for the first time, a dedicated semi-supervised action recognition framework that not only achieves state-of-the-art performance for fine-grained action classes but also performs comparably or better in standard coarse-grained action recognition. Categorically, our method is a pseudo-label-based technique building upon existing videoSSL representations. Our method introduces a novel approach for pseudo-label generation using temporal-alignability-verification-based decisions, which provides a fresh perspective in the semi-supervised action recognition domain. Additionally, our method demonstrates increased robustness to open-world problems, a dimension not previously explored in semi-supervised action recognition. This robustness further distinguishes our approach from the constrained focus of prior work.

\noindent\textbf{Fine-grained Video understanding} There is another line of work that specifically focuses on intra-video dynamics for learning class-agnostic downstream tasks like action-phase classification, Kendall's tau~\cite{tcc}, Aligned Phase Agreement~\cite{dave2024sync} etc. Some recent works have demonstrated the learning of powerful fine-grained intra-video representations in a weakly-supervised manner ~\cite{tcc,bansal2022my,lav,gta} and even on unlabeled data utilizing intra-video self-supervised techniques like ~\cite{carl,vsp, misra2016shuffle,tcn, dave2024no}.

Interestingly, some of these works use alignment-based training objectives to resolve class-agnostic tasks~\cite{egoexo, lav, gta, d3tw}. However, they strictly assume that videos are `alignable' (from the same action class) and do not explore leveraging the alignment property across video classes to learn data-efficient fine-grained action classification. In contrast to the typical `alignment' objective, we opt for an `alignability' objective, where we decide if an unlabeled video belongs to a fine-grained class based on how well it aligns with the limited labeled samples. To the best of our knowledge, we are the first to leverage `alignability'-based intra-video representations in the video-level action recognition task in a semi-supervised setup. For a detailed comparison with prior work, refer \supp{Supp. Sec.F}.

\section{Method}
In semi-supervised action recognition, a limited labeled set \(\mathbb{D}_{l} = \{(\mathbf{v}^{(i)}, \mathbf{y}^{(i)})\}_{i=1}^{N_{l}}\) comprising video instances and their associated action labels is employed alongside a significantly larger unlabeled set \(\mathbb{D}_{u} = \{\mathbf{v}^{(i)}\}_{i=1}^{N_{u}}\). The goal is to leverage both these sets to enhance the performance of action recognition.

Our framework, FinePseudo, is a pseudo-labeling-based co-training approach, as depicted in the schematic diagram in Fig.~\ref{fig:colab_pl}. It mainly consists of two branches: (1) Action encoder $f_E$ responsible for learning high-level video-semantics features such as actions and (2) Auxiliary alignability-encoder $f_A$ which is a frame-wise video encoder - video transformer network (VTN)~\cite{vtn}, to focus on learning the low-level intra-video representations stemming from action phases. 

In this section, we provide more method details of our framework.
First, $f_A$ is trained through alignability-verification-based metric learning from the labeled data (Sec.\ref{sec:metric}). Then, for pseudo-labeling from the unlabeled data, the trained $f_A$ is utilized to provide learned alignability scores for each class, which are passed through a non-parametric classifier to obtain classwise predictions. This alignability-based prediction from $f_A$ is combined with the prediction from the regular video encoder $f_E$ to obtain a collaborative pseudo-label, which is used for the self-training process (Sec.\ref{sec:pl}). A complete algorithm for our FinePseudo training is provided in Sec.~\ref{sec:algo}.

\begin{figure}[t]
    \centering
    \includegraphics[width=\textwidth]{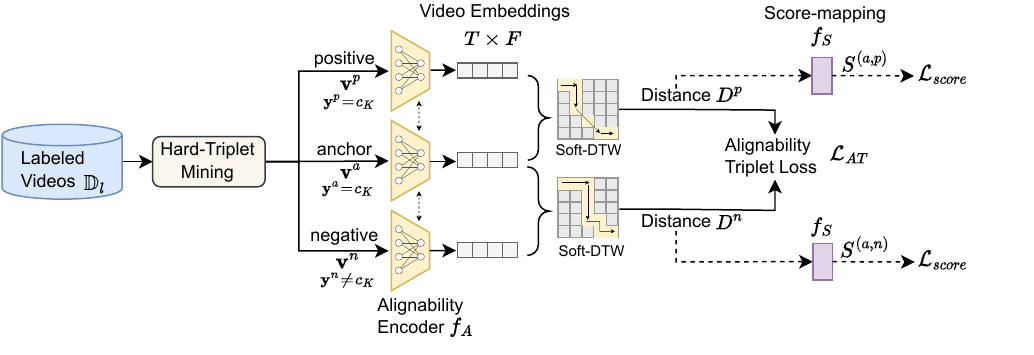}
    \caption{\textbf{Alignability-Verification based Metric Learning} is proposed to is proposed to decide how well two video instances are alignable and produce an `alignability score' for effective learning from a limited labeled set $\mathbb{D}_{l}$. Our approach employs a triplet loss ($\mathcal{L}_{AT}$), considering videos from identical action classes as positive and those from different classes as negative. We selectively mine hard-negatives from the sampled minibatch based on alignment distance, presenting a challenging learning task for the model $f_A$. Additionally, we incorporate a matching loss $\mathcal{L}_{score}$ to quantify the alignment between videos, serving as a verification task to determine whether a video pair belongs to the same class (i.e. alignable or target label = 1) or different classes (i.e. non-alignable or target label = 0). Further details are provided in Sec.~\ref{sec:metric}.}
    \label{fig:enter-label}
\end{figure}

\subsection{Alignability-Verification based Metric Learning}
\label{sec:metric}

The underlying hypothesis is that video instances from the same action class are more alignable compared to those from different classes (as seen in Fig.~\ref{fig:teaser_a}(c)). The objective of this training stage is to solve the alignability verification task, which determines how well two videos are alignable. This knowledge is critical for producing a reliable `learnable alignability score' for a pair of labeled and unlabeled video instances, subsequently aiding in the improvement of pseudo-label quality through a non-parametric classifier within the self-training paradigm.

In our approach, class labels are utilized in learning the alignability-verification task which is a binary classification problem, distinct from the regular multi-class classification setting~\cite{otam}. This strategy encourages the network to focus on differentiating pairs from the same or different classes based on their alignment distance, promoting the learning of more class-agnostic intra-video features. %

\noindent\textbf{Alignment Cost:} Consider a pair of videos \( U \) and \( V \), each with \( T \) frames. To compute the alignment cost between these videos, they are processed through the \( f_A \) network, yielding frame-wise video embeddings \( \mathbf{u}\) and \( \mathbf{v}\), each of shape \( T \times F \), where \( F \) represents the output feature size of \( f_A \). An element-wise cost matrix \( \mathbb{C} \in \mathbb{R}^{T\times T} \) is constructed, with each element computed using the cosine distance: \( \mathbb{C}(i,j) = h(\mathbf{u}(i), \mathbf{v}(j)) \). To identify the optimal alignment path, softDTW~\cite{softdtw}, a differentiable variant of the dynamic time warping algorithm~\cite{dtw}, is utilized. The softDTW distance, \( D(\mathbf{u}, \mathbf{v}) \), is then calculated using the following recursive formula:
\begin{equation}
    D(\mathbf{u}, \mathbf{v}) = \mathbb{C}(i, j) + \gamma\text{-smooth-min}(\Pi_{\text{cost}}(i, j))
    \label{eq:softdtw}
\end{equation}
The function \( \gamma\text{-smooth-min} \) performs a differentiable minimum operation of the possible costs \( \Pi_{\text{cost}}(i, j) \) from the point \( (i, j) \) along the paths $(i,j-1)$, $(i-1,j)$, and $(i-1,j-1)$. Now, we utilize this alignment-cost as the distance to build our metric learning training objective.

\noindent\textbf{Alignability-verification Triplet loss:} 
For a labeled instance $\mathbf{V}^{(i)}$ of class $y^{(i)}=c_K$, we select another instance $\mathbf{V}^{(j)}$ of the same class as positive and an instance $\mathbf{V}^{(k)}$ from a different class as negative. After obtaining the video-embeddings, the alignability-based triplet loss is computed as follows:
\begin{equation}
\mathcal{L}_{AT} = \sum_{i=1}^{N} \left[ D(\mathbf{v}^{(i)}, \mathbf{v}^{(j)}) - D(\mathbf{v}^{(i)}, \mathbf{v}^{(k)}) + \textit{m} \right]
\label{eq:at}
\end{equation}
where $D$ denotes the softDTW distance, \textit{m} is the margin of the triplet loss, and $N$ is the number of samples in the mini-batch $B$. Hard-negative mining is employed from the same mini-batch $B$ for constructing these triplets, with further analysis in \supp{Supp. Sec. C}.

\noindent\textbf{Learnable Alignability Score:} Finally, to determine the alignability of video pairs based on their alignment cost, we propose a normalized scale ranging from 0 (not alignable) to 1 (fully alignable). The computed distance $D$ is mapped through a non-linear scaling function $f_S$ and passed through a sigmoid activation ($\varsigma$) to yield a learnable Alignability-score $S$ between any sequence embeddings $\mathbf{u}$ and $\mathbf{v}$. 
\begin{equation}
    S(\mathbf{u}, \mathbf{v}) = \varsigma(f_S(D(\mathbf{u}, \mathbf{v})))
\end{equation}
To train this scaling function, a binary cross-entropy loss function is employed:
\begin{equation}
    \mathcal{L}_{Score} = -[ y_A \log(S(\mathbf{u}, \mathbf{v})) + (1 - y_A) \log(1 - S(\mathbf{u}, \mathbf{v})) ]
\end{equation}
Where $y_A$ label is assigned 0 for the negative pair and 1 for the positive pair. The overall training objective for our alignability-verification-based metric learning can be expressed as:
\begin{equation}
    \mathcal{L}_{AV} = \mathcal{L}_{AT} + \omega\mathcal{L}_{Score}
    \label{eq:av_loss}
\end{equation}
where, $\omega$ hyperparameter is the relative weighting factor.

While the alignability encoder ($f_A$) is trained through the alignability-verification training from the labeled set $\mathbb{D}_l$, the action encoder ($f_E$) is trained through regular cross-entropy loss as shown in the equation below:
\begin{equation}
    \mathcal{L}^{(i)}_{CE} = -\sum_{c=1}^{N_c} \mathbf{y}^{(i)}_{c}\log \mathbf{p}^{(i)}_{c}
    \label{eq:cross_entropy}
\end{equation}

\noindent Where $N_c$ is the number of classes, ${y}^{(i)}_{c}$ is the ground-truth class and $\mathbf{p}_{c}$ is the classwise prediction by classification head of $f_E$.

\subsection{Collaborative Pseudo-Labeling}
\label{sec:pl}

Once both action encoder $f_E$ and alignability encoder $f_A$ is trained with the $\mathbb{D}_l$, they are utilized to generate pseudo-labels for the videos of unlabeled set $\mathbb{D}_u$. 
Before we start pseudo-labeling, we first construct a set $\mathbb{A}$ by obtaining embedding of video of $D_l$ by passing it through encoder $f_A$.  This process is formalized as $\mathbb{A} = \{ f_A(\mathbf{v}^{(i)}) \}_{i=1}^{N_{l}}$

\begin{figure}[t]
    \centering
    \includegraphics[width=\textwidth]{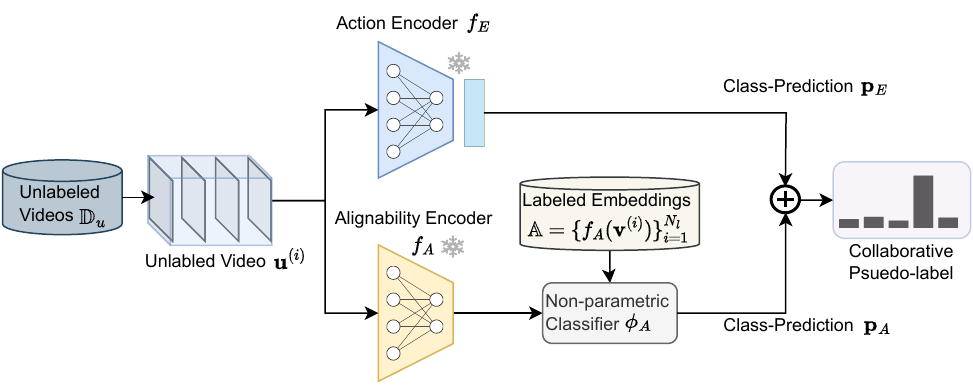}
    \caption{\textbf{Collaborative Pseudo-labeling:} The unlabeled instance $\mathbf{u}^{(i)}$ undergoes processing by both video encoders ($f_E$ and $f_A$). For the Action Encoder $f_E$, its prediction ($\mathbf{p}_E$) is derived via its classification head. For the Alignability Encoder $f_A$, the embedding of $\mathbf{u}^{(i)}$ computes class-wise alignability scores against a gallery of labeled embeddings $\mathbb{A}$. These scores are then used to generate a class-wise prediction $\mathbf{p}_{A}$ using the non-parametric classifier $\phi_A$. As these predictions stem from distinct supervisory signals—$\mathbf{p}_E$ from video-level and $\mathbf{p}_A$ from alignability-based supervision—they offer complementary insights, resulting in a refined collaborative pseudo-label.}
    \label{fig:colab_pl}
\end{figure}

For an unlabeled video $U\in\mathbb{D}_u$, its embedding $\mathbf{u}$ is obtained by passing it through the alignability encoder $f_A$. The alignability score for each class $c$ in the labeled dataset is computed by randomly sampling $\rho$ embeddings from $\mathbb{A}$ corresponding to class $c$, denoted as $\mathbb{A}_c^\rho$. The average alignability score $\bar{S}_c$ for class $c$ is calculated as:
\begin{equation}
\bar{S}_c = \frac{1}{\rho} \sum_{\mathbf{a} \in \mathbb{A}_c^\rho} S(\mathbf{u}, \mathbf{a})
\label{eq:alignability_score}
\end{equation}
For computing the class prediction $\mathbf{p}_A$ for the unlabeled video $U$ using the softmax function with a temperature parameter $\tau$. This function is applied to the alignability scores, yielding the class prediction as:
\begin{equation}
\mathbf{p}_A(c) = \frac{\exp(\bar{S}_c / \tau)}{\sum_{j} \exp(\bar{S}_j / \tau)}
\label{eq:class_prediction}
\end{equation}
The denominator in Eq.~\ref{eq:class_prediction} sums over all classes $j$ in $\mathbb{D}_{l}$, producing a probability distribution over the classes and indicating the predicted likelihood of the unlabeled video $U$ belonging to each class. Since there is no parameter involved in getting the prediction $\mathbf{p}_A$ we can call it non-parametric classifier $\phi_A$ of the alignability encoder. 

The same unlabeled video U is passed through $f_E$ and its classifier head to obtain its class prediction $\mathbf{p}_E$. 
The overall final prediction $\mathbf{p}$ for a video $U$ is obtained by adding the predictions from both classifiers: $\mathbf{p} = \mathbf{p}_A + \mathbf{p}_E$.
We apply a confidence threshold $\theta$ to each prediction $\mathbf{p}$. If the highest confidence score in the prediction $\mathbf{p}$ exceeds the threshold $\theta$, the sample is considered for generating a hard pseudo-label; otherwise, the sample is discarded. 
In this way, we achieve refined pseudo-labels and they are used for the next iteration of labeled training for both $f_A$ and $f_E$.

\subsection{Algorithm}
\label{sec:algo}
\begin{algorithm}[t]
\textbf{Inputs}:

          \hspace*{0.2cm}\textit{Datasets:} $\mathbb{D}_{u}$, $\mathbb{D}_{l}$
         
          \hspace*{0.2cm}\textit{\#Epochs:} $max\_epoch\_ssl$, $max\_epoch\_labeled$, $max\_iter$, $max\_epoch\_st$
         
          \hspace*{0.2cm}\textit{Learning Rates:} $\alpha_A$, $\alpha_E$
         
          \hspace*{0.2cm}\textit{Hyperparameters:} Confidence threshold $\theta$

\textbf{Output}: Action Encoder model $\theta_E$
\SetAlgoLined
\hrule 
\vspace{1mm}%
SSL Pretraining on Unlabeled Set $\mathbb{D}_u$:

\For{$e_0 \gets 1$ \KwTo $max\_epoch\_ssl$}{
    $\theta_A \gets \theta_A - \alpha_A \nabla \mathcal{L}_{GITDL}(\theta_A)$ (Refer ~\supp{Supp. Eq.~1})
}
\hrule 
\vspace{1mm}%
Training from the Labeled Set $\mathbb{D}_l$:

\For{$e_0 \gets 1$ \KwTo $max\_epoch\_labeled$}{
    $\theta_E \gets \theta_E - \alpha_E \nabla \mathcal{L}_{CE}(\theta_E)$ (Refer Eq.~\ref{eq:cross_entropy})}
\For{$e_0 \gets 1$ \KwTo $max\_epoch\_labeled$}{
    $\theta_A \gets \theta_A - \alpha_A \nabla \mathcal{L}_{AV}(\theta_A)$ (Refer Eq.~\ref{eq:av_loss})
}
\hrule 
\vspace{1mm}
Self-Training through Collaborative Pseudo-Labeling:

\For{$iter \gets 1$ \KwTo $max\_iter$}{
    \For{each sample in $\mathbb{D}_u$}{
        Obtain combined class-prediction 
            \hspace*{0.2cm}$\mathbf{p} = \text{avg}(\mathbf{p}_A, \mathbf{p}_E)$ \\
            \hspace*{0.2cm} Predicted class $\hat{\mathbf{y}}$ \\
        \If{confidence of $\hat{\mathbf{y}} > \theta$}{
            Add (sample, predicted label $\hat{\mathbf{y}}$) to $\mathbb{D}_l$
        }
    }
    \For{$epoch_0 \gets 1$ \KwTo $max\_epoch\_st$}{
        $\theta_E \gets \theta_E - \alpha_E \nabla \mathcal{L}_{CE}(\theta_E)$ \\
        $\theta_A \gets \theta_A - \alpha_A \nabla \mathcal{L}_{AV}(\theta_A)$
    }
}
\caption{FinePseudo Training Algorithm}
\label{alg:finepseudo_algo}
\end{algorithm}

Let's consider the action encoder model $f_{E}$ and the alignability model $f_{A}$, parameterized by $\theta_{E}$ and $\theta_{A}$, respectively. In our semi-supervised training framework, firstly we employ our novel GITDL-based self-supervised pretraining (Details in ~\supp{Supp. Sec. E}) on the unlabeled dataset $\mathbb{D}_{u}$ to learn frame-wise video representations focusing on intra-video dynamics, and secondly, leveraging both labeled $\mathbb{D}_{l}$ and pseudo-labeled data in a collaborative self-training process. These steps are put together in Algorithm~\ref{alg:finepseudo_algo}, which outlines the complete process for our \textit{FinePseudo} framework for semi-supervised action recognition.

\section{Experiments}
\subsection{Datasets and Metrics}
\noindent \textbf{Diving48}~\cite{diving} is a fine-grained dataset on competitive diving, with 48 distinct patterns across roughly 18k videos. Each class underscores the intricacies of a diver's movements, stressing the need for detailed temporal analysis to capture subtle differences in takeoff, flight, and entry phases.

\noindent \textbf{FineGym}~\cite{shao2020finegym} is a large-scale, fine-grained action recognition dataset that provides hierarchical annotations for four different gymnastic events: Vault, Floor Exercise, and Balance Beam. 
It comprises two main splits: FineGym99 with 99 actions from ~29k videos, and FineGym288 with 288 actions from ~32k videos.

\noindent \textbf{FineDiving}~\cite{xu2022finediving} dataset comprises diverse diving events, covering 52 action classes across 23 difficulty degrees.

\noindent\textbf{Kinetics400}~\cite{kinetics} encompasses 400 human action classes across approximately 260k videos sourced from YouTube. 

\noindent \textbf{Something-SomethingV2}~\cite{goyal2017something} is another large dataset with clips that are object class agnostic, focusing on a wide range of 174 hand-object interactions.

\noindent For further dataset details, refer \supp{Supp. Sec. A}.

\noindent\textbf{Evaluation Metric}: Following standard protocols in prior work~\cite{timebalance, semi_cmpl}, we evaluate 3 independent label splits and report the mean Top-1 accuracy.

\noindent For implementation details, refer \supp{Supp. Sec. B}

\begin{table}[t]
\renewcommand{\arraystretch}{1.05}

\centering
\arrayrulecolor[rgb]{0.753,0.753,0.753}
\caption{Comparison with state-of-the-art semi-supervised methods on Fine-grained Action recognition datasets under various \% of labeled data setting. Highlighted \bestresult{Red} shows the best results and \secondbest{Blue} shows second best results. All results are reported on R2plus1D-18 utilizing the exact same amount of training data.}
\begingroup
\resizebox{\linewidth}{!}{
\setlength{\tabcolsep}{1.5pt}
\begin{tabular}{l!{\color{black}\vrule}c|c|c!{\color{black}\vrule}c|c|c!{\color{black}\vrule}c|c|c!{\color{black}\vrule}c|c|c} 
\arrayrulecolor{black}\hline

\hline

\hline\\[-3mm]

\multirow{2}{*}{\textbf{Method}} & \multicolumn{3}{c!{\color{black}\vrule}}{\textbf{Diving48}} & \multicolumn{3}{c!{\color{black}\vrule}}{\textbf{FineGym99}} & \multicolumn{3}{c!{\color{black}\vrule}}{\textbf{FineGym288}} & \multicolumn{3}{c}{\textbf{FineDiving}} \\ 
\cline{2-13}
                                 & \textbf{5\%} & \textbf{10\%} & \textbf{20\%} & \textbf{5\%} & \textbf{10\%} & \textbf{20\%} & \textbf{5\%} & \textbf{10\%} & \textbf{20\%} & \textbf{5\%} & \textbf{10\%} & \textbf{20\%} \\ 
\hline
TCLR~\venueTT{CVIU'22\cite{tclr}}                              & 14.3 & 33.1 & 53.7 & 43.2 & 64.2 & 74.9 & 36.0 & 56.8 & 67.2 & 23.2 & 42.3 & 65.2 \\
VidMoCo~\venueTT{CVPR'21\cite{videomoco}}                        & 12.6 & 31.4 & 52.5 & 41.6 & 62.8 & 73.8 & 34.2 & 55.8 & 66.8 & 21.9 & 40.6 & 64.8 \\
GDT~\venueTT{ICCV'21\cite{gdt}                   }           & 12.2 & 31.7 & 51.8 & 42.0 & 62.0 & 73.3 & 35.3 & 56.0 & 66.6 & 21.2 & 40.9 & 64.3 \\
AVID~\venueTT{CVPR'21\cite{avidcma}}                         & 10.0 & 30.4 & 51.5 & 40.3 & 60.3 & 72.7 & 32.5 & 55.6 & 64.5 & 20.6 & 39.6 & 62.7 \\ 
\arrayrulecolor[rgb]{0.753,0.753,0.753}
RSPNet~\venueTT{AAAI'21\cite{chen2021rspnet}}                           & 14.0 & 33.0 & 53.7 & 43.4 & 64.0 & 75.2 & 36.8 & 56.4 & 67.1 & 23.0 & 42.5 & 65.1 \\
 \arrayrulecolor{black} \midrule
PL~\venueTT{ICML'13\cite{lee2013pseudo}}                                & 14.4 & 33.4 & 54.0 & 43.2 & 64.4 & 75.1 & 34.9 & 55.5 & 67.1 & 23.5 & 42.0 & 66.1 \\
UPS~\venueTT{ICLR'21\cite{ups}}                               & 14.6 & 33.6 & 54.1 & - & - & - & - & - & - & - & - & - \\
ActorCM~\venueTT{CVIU'22\cite{semi_actorcutmix}}                                & 14.7 & 33.8 & 54.7 & 43.8 & 65.0 & 75.9 & 36.5 & 56.9 & 67.7 & - & - & - \\
TG-FM~\venueTT{CVPR'21\cite{semi_tgfixmatch}}                       & \secondbest{16.0} & \secondbest{33.8} & 54.4 & 44.1 & 64.9 & 75.7 & 36.9 & 56.6 & 67.6 & - & - & - \\
TimeBal~\venueTT{CVPR'23\cite{timebalance}}                       & 15.8 & 33.7 & \secondbest{56.3} & \secondbest{44.4} & \secondbest{65.9} & \secondbest{76.1} & \secondbest{37.3} & \secondbest{57.8} & \secondbest{68.6} & \secondbest{25.1} & \secondbest{43.9} & \secondbest{67.5} \\
\rowcolor[rgb]{0.784,0.902,0.976}Ours (\textit{FinePseudo})                      & \bestresult{20.9} & \bestresult{37.6} & \bestresult{60.4} & \bestresult{49.2} & \bestresult{69.9} & \bestresult{80.0} & \bestresult{41.7} & \bestresult{62.5} & \bestresult{73.4} & \bestresult{28.4} & \bestresult{46.8} & \bestresult{71.9} \\
\arrayrulecolor{black}\bottomrule
\end{tabular}
}
\endgroup
\label{table:bigFG}
\end{table}

\subsection{Evaluation on Fine-grained datasets}
In order to maintain comparability across methods, we utilize the R2plus1D-18 network. 
We compare various baselines such as video self-supervised methods~\cite{tclr,videomoco,avid,gdt,chen2021rspnet}, classical semi-supervised learning baselines~\cite{Lee2013PseudoLabelT,ups}, and state-of-the-art video semi-supervised methods~\cite{semi_actorcutmix,semi_tgfixmatch, timebalance} in Table~\ref{table:bigFG}. In the first section of Table~\ref{table:bigFG}, we study video self-supervised methods by taking their publicly available Kinetics400 self-supervised weights and fine-tuning them for the fine-grained action recognition task under limited labeled data. We observe that the methods~\cite{tclr} and \cite{chen2021rspnet}, which explicitly promote temporal distinctiveness, perform better than other video self-supervised methods. 

Based on this observation, we use the best-performing SSL weights~\cite{tclr} for \textbf{all semi-supervised methods }in the second part of Table~\ref{table:bigFG}. Firstly, we note that classical semi-supervised baselines, namely PL and UPS, do not perform as well compared to video semi-supervised methods. Our method consistently outperforms all prior methods by an absolute \textbf{4-5\%} in terms of top-1 accuracy.

\noindent\textbf{Evaluating with Transformer architecture} Since the AIM-ViTB architecture~\cite{aim_chen} achieves state-of-the-art performance on the Diving48 dataset in a fully-supervised setting, we find it interesting to base our comparisons. In this architecture, the ViT-B backbone~\cite{vit} is kept frozen and initialized with the CLIP~\cite{clip} visual encoder, and spatio-temporal adaptor layers are trained.

\setlength{\intextsep}{2pt} %
\begin{wraptable}{r}{0.45\textwidth} %
\renewcommand{\arraystretch}{1.1}
\arrayrulecolor{black}
\centering
\caption{Results with AIM model on Diving48 dataset}
\begingroup
\resizebox{\linewidth}{!}{
\setlength{\tabcolsep}{3.5pt}
\begin{tabular}{l|ccc} 
\arrayrulecolor{black}\hline

\hline

\hline\\[-3mm]
\textbf{Method} & \textbf{5\%} & \textbf{10\%} & \textbf{20\%}  \\ 
\hline
Supervised & 37.28 & 55.33 & 75.36  \\
PL~\cite{Lee2013PseudoLabelT} & 37.33 & 55.40 & 75.42  \\
UPS~\cite{ups} & 37.70 & 55.61 & 75.56  \\
SVFormer~\cite{xing2023svformer} & 38.00 & \secondbest{56.02} & \secondbest{76.20}  \\
TimeBal~\cite{timebalance} & \secondbest{38.12} & 55.80 & 76.01  \\
\rowcolor[rgb]{0.784,0.902,0.976}Ours & \bestresult{43.02} & \bestresult{60.79} & \bestresult{80.02}  \\
\arrayrulecolor{black}\bottomrule
\end{tabular}
}
\label{table:aim}
\endgroup
\end{wraptable}

Firstly, using this architecture (Table~\ref{table:aim}), we observe a significant improvement compared to Table~\ref{table:bigFG}. 
Next, examining the results of recent semi-supervised methods~\cite{xing2023svformer, timebalance}, it becomes evident that token-mix augmentations from ~\cite{xing2023svformer} are not as effective in fine-grained datasets as in coarse-grained ones. Similarly, videoSSL-based semi-supervised methods like ~\cite{timebalance} also underperform due to the ineffectiveness of some components like temporally-invariant teacher in fine-grained datasets. 
Our method achieves a clear improvement of \textbf{4-5\%}, \textit{demonstrating its potential to further enhance the strong foundational model pretraining for fine-grained action recognition in a limited labeled setting}.

\subsection{Evaluation on Coarse-grained action datasets}
Although the focus of our work is on the evaluation of fine-grained actions, we also evaluate coarse-grained action datasets as shown in Table~\ref{table:bigExisting}. For comparability, results are presented using two backbones: 3D-ResNet18 and 3D-ResNet50~\cite{slowfast}, with an input resolution of $224\times224$ and $8$-frame clips. Our learnable-alignability score-based approach shows favorable or slightly improved performance over prior best methods across both backbones. This demonstrates that our approach, not reliant on a strict alignment criterion, generalizes well for generic coarse-grained human actions and is not confined to fine-grained actions.

\begin{table}[t]
\renewcommand{\arraystretch}{1.1}

\centering
\caption{Results on standard Coarse-grained Action recognition datasets at various \% of labeled set. Highlighted \bestresult{Red} shows the best and \secondbest{Blue} shows second best results. }

\arrayrulecolor[rgb]{0.753,0.753,0.753}
\begingroup
\resizebox{\linewidth}{!}{
\setlength{\tabcolsep}{1.4pt}
\begin{tabular}{llccc!{\color{black}\vrule}c|c|c!{\color{black}\vrule}c|c|c} 

\arrayrulecolor{black}\hline

\hline

\hline\\[-3mm]

\multirow{2}{*}{\textbf{Method}} & \multirow{2}{*}{\textbf{Backbone}} & \multirow{2}{*}{\begin{tabular}[c]{@{}c@{}}\textbf{Params}\\\textbf{(M)}\end{tabular}} & \multirow{2}{*}{\begin{tabular}[c]{@{}c@{}}\textbf{ImgNet}\\\textbf{init?}\end{tabular}} & \multirow{2}{*}{\textbf{\#F}} & \multicolumn{3}{c!{\color{black}\vrule}}{\textbf{Kinetics400}} & \multicolumn{3}{c}{\textbf{S. SomethingV2}}   \\ 
\cline{6-11}
                                 &                                    &                                                                                        &                                                                                          &                               & \textbf{1\%} & \textbf{5\%} & \textbf{10\%}                    & \textbf{1\%} & \textbf{5\%} & \textbf{10\%}  \\ 
\hline
MT~\venueTT{NeuRIPS'17\cite{tarvainen2017mean}}                      & TSM-ResNet18                       & 13                                                                                     & \xmark                                                                                      & 8                             & 6.8         & 23.0        & -                                & 7.3         & 20.2        & 30.2          \\
S4L~\venueTT{ICCV'19\cite{s4l}}                              & TSM-ResNet18                       & 13                                                                                     & \xmark                                                                                      & 8                             & 6.3         & 23.3        & -                                & 7.2         & 18.6        & 26.0          \\
MM~\venueTT{NeuRIPS'19\cite{NIPS2019_8749_MixMatch}}                         & TSM-ResNet18                       & 13                                                                                     & \xmark                                                                                      & 8                             & 7.0         & 21.9        & -                                & 7.5         & 18.6        & 25.8          \\
FM~\venueTT{NeuRIPS'20\cite{fixmatch}}                         & TSM-ResNet18                       & 13                                                                                     & \xmark                                                                                      & 8                             & 6.4         & 25.7        & -                                & 6.0         & 21.7        & 33.4          \\
TCL~\venueTT{CVPR'21\cite{semi_tcl}}                              & TSM-ResNet18                       & 13                                                                                     & \xmark                                                                                      & 8                             & 11.6        & \secondbest{31.9}        & -                                & \secondbest{9.9}         & \secondbest{31.0}        & \secondbest{41.6}          \\
TG-FM~\venueTT{CVPR'21\cite{semi_tgfixmatch}}                      & 3D-ResNet18                        & 13.5                                                                                   & \xmark                                                                                      & 8                             & 9.8         & -            & 43.8                            & -            & -            & -              \\
MvPL~\venueTT{ICCV'21\cite{semi_mvpl}}                             & 3D-ResNet18                        & 13.5                                                                                   & \xmark                                                                                      & 8                             & 5.0         & -            & 36.9                            & -            & -            & -              \\
CMPL~\venueTT{CVPR'22\cite{semi_cmpl}}                             & 3D-ResNet18                        & 13.5                                                                                   & \xmark                                                                                      & 8                             & 16.5        & -            & 53.7                            & -            & -            & -              \\
TimeBal~\venueTT{CVPR'23\cite{timebalance}}                      & 3D-ResNet18                        & 13.5                                                                                   & \xmark                                                                                      & 8                             & \secondbest{17.1}        & -            & \secondbest{54.9}                            & -            & -            & -              \\
\rowcolor[rgb]{0.784,0.902,0.976}Ours (\textit{FinePseudo})                     & 3D-ResNet18                        & 13.5                                                                                   & \xmark                                                                                      & 8                             & \bestresult{18.6}        & \bestresult{43.2}            & \bestresult{56.1}                                & \bestresult{13.1}            & \bestresult{34.3}            & \bestresult{45.4}          \\ 
\midrule
FM~\venueTT{NeuRIPS'20\cite{fixmatch}}                         & SlowFast-R50                       & 60                                                                                     & \xmark                                                                                      & 8                             & 10.1        & -            & 49.4                            & 6.5         & 25.3        & 37.4          \\
MvPL~\venueTT{ICCV'21\cite{semi_mvpl}}                             & 3D-ResNet50                        & 31.8                                                                                   & \xmark                                                                                      & 8                             & 17.0        & -            & 58.2                            & -            & -            & -              \\
CMPL~\venueTT{CVPR'22\cite{semi_cmpl}}                             & 3D-ResNet50                        & 31.8                                                                                   & \xmark                                                                                      & 8                             & 17.6        & -            & 58.4                            & -            & -            & -              \\
TimeBal~\venueTT{CVPR'23\cite{timebalance}}                      & 3D-ResNet50                        & 31.8                                                                                   & \xmark                                                                                      & 8                             & \secondbest{19.6}        & -            & \secondbest{61.2}                            & -            & -            & -              \\
SVFormer~\venueTT{CVPR'23\cite{xing2023svformer}}                         & T.Former(ViT-S)                  & 30.7                                                                                   & \xmark                                                                                      & 16                            & 17.2        & \secondbest{42.3}        & 58.1                            & \secondbest{9.9}         & \secondbest{31.7}        & \secondbest{42.9}          \\
\rowcolor[rgb]{0.784,0.902,0.976}Ours (\textit{FinePseudo})                     & 3D-ResNet50                        & 31.8                                                                                   & \xmark                                                                                      & 8                             & \bestresult{21.4}        & \bestresult{47.5}            & \bestresult{62.6}                                & \bestresult{13.4}            & \bestresult{34.7}            & \bestresult{46.1}          \\
\arrayrulecolor{black}\bottomrule
\end{tabular}
}
\endgroup
\label{table:bigExisting}
\end{table}

\subsection{Evaluation on Open-World setting}
\setlength{\intextsep}{2pt} %
\begin{wraptable}{r}{0.4\textwidth}
\arrayrulecolor{black}
\centering
\caption{Results with open-world setting on Diving48 dataset. All models are R2plus1D-18.}
\begingroup
\resizebox{\linewidth}{!}{
\setlength{\tabcolsep}{4pt}
\begin{tabular}{l|cc} %
\arrayrulecolor{black}\hline

\hline

\hline\\[-3mm]

\textbf{Method} & \textbf{10\%} & \textbf{20\%}  \\ %
\hline
Supervised      & 39.60         & 50.23          \\
Pseudo-labeling & 38.29         & 49.41          \\
UPS~\cite{ups}  & 38.93         & 49.56          \\
TimeBalance~\cite{timebalance} & \secondbest{39.90} & \secondbest{50.88} \\
\rowcolor[rgb]{0.784,0.902,0.976}Ours            & \bestresult{42.21} & \bestresult{55.37} \\
\arrayrulecolor{black}\bottomrule
\end{tabular}
}
\endgroup
\label{table:openworld}
\end{wraptable}

In previous evaluations, it was assumed that the unlabeled data belonged to one of the classes in the labeled set. However, in practical scenarios, an unlabeled sample could originate from any \textit{novel} (unknown) action class. Refer to \supp{Supp. Sec. E} for more details about this protocol.

To explore the open-world setting, we utilize the Diving48 dataset, where 40 classes are randomly selected as \textit{known} classes and the remaining 8 classes are designated as \textit{novel} classes. For this protocol, we consider the R2plus1D-18 model with SSL initialization from Kinetics400 from ~\cite{tclr}, and the results are reported in Table~\ref{table:openworld}. The supervised baseline, which only utilizes the labeled data from the 40 classes, is established for comparison. The regular pseudo-label setting degrades the performance of the supervised baseline, as the novel unlabeled samples introduce noise during self-training. The prior best semi-supervised method~\cite{timebalance} also fails to show noticeable improvement over the supervised baseline, as its teacher model categorizes the unlabeled sample into one of the known classes before distillation to a student. In contrast, our approach, with its non-parametric classification in PL generation, effectively filters out unknown classes based on low alignability scores, thereby achieving improvement over other methods. \noindent For additional results, refer \supp{Supp. Sec. D}.

\subsection{Ablation Study}
\label{sec:ablation}
We demonstrate the ablation experiments on Diving48 dataset with R2plus1D-18 network by default. Additional ablations and detail in \supp{Supp. Sec. C}.

\begin{table}[t]
\renewcommand{\arraystretch}{1}

\centering
\caption{Ablation with different components of framework}

\begingroup
\setlength{\tabcolsep}{3pt}

\begin{tabular}{cc|c!{\color[rgb]{0.753,0.753,0.753}\vrule}cc|cc} 
\arrayrulecolor{black}\hline

\hline

\hline\\[-3mm]
    & \multirow{3}{*}{\begin{tabular}[c]{@{}c@{}}\textbf{Action}\\\textbf{Encoder} \\\textbf{\textbf{$f_E$}}\end{tabular}} & \multicolumn{3}{c|}{\textbf{Alignability Encoder $f_A$}}                  & \multicolumn{2}{c}{\textbf{Top-1 Accuracy}}                  \\ 
\cline{3-5}
    &                                                                                                                   & \textbf{SSL ($\mathbb{D}_U$)} & \multicolumn{2}{c|}{\textbf{Metric Learning ($\mathbb{D}_L$)}} & \multirow{2}{*}{\textbf{10\%}} & \multirow{2}{*}{\textbf{20\%}}  \\
    &                                                                                                                   & $\mathcal{L}_{GITDL}$ & $\mathcal{L}_{AT}$ & $\mathcal{L}_{Score}$                     &                                &                                 \\ 
\hline
\texttt{(PL)}  & \textcolor[rgb]{0.7,0.7,0.7}{\cmark}                                                                                                                & -                 & -            & -                                   & 33.40                          & 54.00                           \\
\hline
\texttt{(a)}  & \textcolor[rgb]{0.7,0.7,0.7}{\cmark}                                                                                                                & -                 & -            & -                                   & 33.10                          & 53.70                           \\
\texttt{(b)}  & \xmark                                                                                                                & \textcolor[rgb]{0.7,0.7,0.7}{\cmark}                & \textcolor[rgb]{0.7,0.7,0.7}{\cmark}           & \textcolor[rgb]{0.7,0.7,0.7}{\cmark}                                  & 32.82                          & 51.05                           \\
\texttt{(c)}  & \textcolor[rgb]{0.7,0.7,0.7}{\cmark}                                                                                                                & \textcolor[rgb]{0.7,0.7,0.7}{\cmark}                & \xmark           & \xmark                                  & 33.50                          & 53.76                           \\
\texttt{(d)}  & \textcolor[rgb]{0.7,0.7,0.7}{\cmark}                                                                                                                & \textcolor[rgb]{0.7,0.7,0.7}{\cmark}                & \textcolor[rgb]{0.7,0.7,0.7}{\cmark}           & \xmark                                  & 33.73                          & 55.67                           \\
\texttt{(e)}  & \textcolor[rgb]{0.7,0.7,0.7}{\cmark}                                                                                                                & \textcolor[rgb]{0.7,0.7,0.7}{\cmark}                & \xmark           & \textcolor[rgb]{0.7,0.7,0.7}{\cmark}                                  & 36.11                          & 59.32                           \\
\texttt{(f)}  & \textcolor[rgb]{0.7,0.7,0.7}{\cmark}                                                                                                                & \xmark                & \textcolor[rgb]{0.7,0.7,0.7}{\cmark}           & \textcolor[rgb]{0.7,0.7,0.7}{\cmark}                                  & 35.23                          & 58.64                           \\
\midrule
\texttt{(g)}  & \textcolor[rgb]{0.7,0.7,0.7}{\cmark}                                                                                                                & \textcolor[rgb]{0.7,0.7,0.7}{\cmark}                & \textcolor[rgb]{0.7,0.7,0.7}{\cmark}           & \textcolor[rgb]{0.7,0.7,0.7}{\cmark}                                  & \textbf{37.64}                          & \textbf{60.40}                           \\
\bottomrule
\end{tabular}
\endgroup
\label{table:abl_framework}
\end{table}

\noindent\textbf{Evaluating Contributions of Training Components:} In Table~\ref{table:abl_framework}, we study the effect of each training step in our framework: SSL pretraining on $\mathbb{D}_u$ and Alignability-based metric learning on $\mathbb{D}_l$. 
\begin{itemize}
    \item When using individual video encoders~\texttt{(Rows a, b)}, $f_E$ performs better than $f_A$, however, it is significantly suboptimal compared to their collaborative use in \texttt{(g)}. \texttt{Row (PL)} shows regular PL baseline~\cite{Lee2013PseudoLabelT} for the $f_E$ which helps only by a small margin. 
    The Alignability-Verification-based metric learning significantly help to improve the capability of recognizing fine-grained actions.(\texttt{Row c vs Row g}). 
    \item  \texttt{Row d,e vs Row g} suggest that both alignability-triplet loss and score loss contribute towards the final performance. Since $\mathcal{L}_{AT}$ provides a more challenging task with hard triplets and margin, it helps significantly compared to the simpler binary classification objective of $\mathcal{L}_{Score}$.
    \item Proposed GITDL self-supervised pretraining for $f_A$ helps 2\% on the final performance (\texttt{Row f vs Row g}).

\end{itemize}

\noindent\textbf{Pseudo-Label Refinement Strategies:} We examine the impact of various pseudo-labeling (PL) strategies on the Diving48 dataset with a limited labeled split, as shown in Table~\ref{table:ablpl}. Alongside the final performance, we also report the number of pseudo-labels (PLs) that surpass the threshold and their accuracy, as determined by comparison with the ground truth in the fully labeled set.

\begin{wraptable}{r}{0.59\textwidth}
\centering
\caption{Psuedo-Label refinement methods.}
\renewcommand{\arraystretch}{1.2}

\arrayrulecolor[rgb]{0.753,0.753,0.753}
\begingroup
\resizebox{\linewidth}{!}{
\setlength{\tabcolsep}{1.2pt}
\begin{tabular}{l!{\color{black}\vrule}c|c!{\color{black}\vrule}c|c} 
\arrayrulecolor{black}\hline

\hline

\hline\\[-3mm]
\multirow{2}{*}{\begin{tabular}[c]{@{}l@{}}\textbf{Pseudo-Labelling(PL)}\\\textbf{Method }\end{tabular}} & \multicolumn{2}{c!{\color{black}\vrule}}{\textbf{PL statistics}} & \multicolumn{2}{c}{\textbf{Results }}                                   \\ 
\arrayrulecolor[rgb]{0.753,0.753,0.753}\cline{2-5}
                                                                                                         & \textbf{Count} & \textbf{Acc.} & \textbf{10\%} & \textbf{20\%}  \\ 
\arrayrulecolor{black}\hline
Regular- Conf. based                                                                                     & 4813           & 85.4          & 33.40         & 53.95         \\
Uncertainty based                                                                                        & 3565           & 87.9          & 33.58         & 54.07         \\ 
\hline
Label verification                                                                                       & 1981           & 97.0          & 37.09         & 59.57         \\
Non-Parametric Classif.                                                                                        & 4558           & 96.4          & \textbf{37.64}& \textbf{60.40}\\
\bottomrule
\end{tabular}
}
\endgroup
\label{table:ablpl}
\vspace{2mm}
\end{wraptable}

In the first section, we explore standard pseudo-labeling methods based on the model's class prediction confidence and uncertainty. We find that incorporating uncertainty with confidence (as shown in the second row) enhances PL accuracy but reduces the quantity of PLs. Because of this reduction, the improved PL accuracy does not translate into a noticeable gain in final performance.

In the third row, we introduce an alignability-score based PL verification strategy. After a class prediction by \( f_E \) clears the confidence-based threshold, we calculate the alignability score for its predicted class. If this score exceeds an alignability-score threshold (set at 0.6), we accept the PL for self-training; otherwise, it is discarded. This alignability-based score verification significantly improves PL accuracy and consequently enhances overall performance.

Finally, in the last row, we present results using our combined class prediction approach, which incorporates a prediction \( \mathbf{p}_A \) obtained through a non-parametric (NP) classifier (as detailed in Sec.~\ref{sec:pl}). This method substantially increases the count of PLs over the verification-based PL approach and improves the overall results.

\section{Conclusion and Future Work}
We present FinePseudo, a novel co-training-based semi-supervised framework tailored for fine-grained action recognition. Our framework effectively utilizes the strengths of a coarse-level video encoder dedicated to high-level action understanding, alongside a frame-wise video encoder focusing at capturing low-level intra-video dynamics, particularly action phases. 
Notably, FinePseudo improves existing state-of-the-art video SSL methods and foundational models when trained for semi-supervised learning for fine-grained action recognition. The efficacy of our collaborative pseudo-labeling process is further validated in open-world semi-supervised scenarios.

For future work, exploring multi-modal temporal-alignability, such as video and audio integration, could enhance the efficiency of semi-supervised action recognition. Additionally, the potential of FinePseudo extends to other video understanding tasks requiring fine-grained temporal understanding, like action quality assessment \etc.

\clearpage  %

\section*{Acknowledgements}
This work was supported in part by the National Science Foundation (NSF) and Center for Smart Streetscapes (CS3) under NSF Cooperative Agreement No. EEC-2133516.

\bibliographystyle{splncs04}
\bibliography{main}
% \end{document}
\clearpage
\appendix

\section*{Supplementary Material Overview}
\begin{itemize}
    \item Section~\ref{suppsec:dataset}: Details of the datasets
    \item Section~\ref{suppsec:implementation}: Implementation details about the hyperparameters and training schedule
    \item Section~\ref{suppsec:ablation}: Additional ablation for our framework
    \item Section~\ref{suppsec:results}: Results on additional splits and tasks.
    \item Section~\ref{suppsec:method}: Supportive algorithm and diagrams
    \item Section~\ref{suppsec:priorwork}: Detailed comparison with related prior work
\end{itemize}

\section{Dataset Details}
\label{suppsec:dataset}
All datasets used in our study are publicly available. We utilize only the action class labels from these datasets.

\noindent\textbf{Diving48}~\cite{diving} includes 48 action classes of diving actions. Each sequence is defined by a combination of takeoff (dive groups), movements in flight (somersaults and/or twists), and entry (dive positions). We utilize the V2 set of annotations, which is a cleaner version.

\begin{figure}[h]
    \centering
    \includegraphics[width=\linewidth, trim={5mm 0mm 0mm 5mm}, clip]{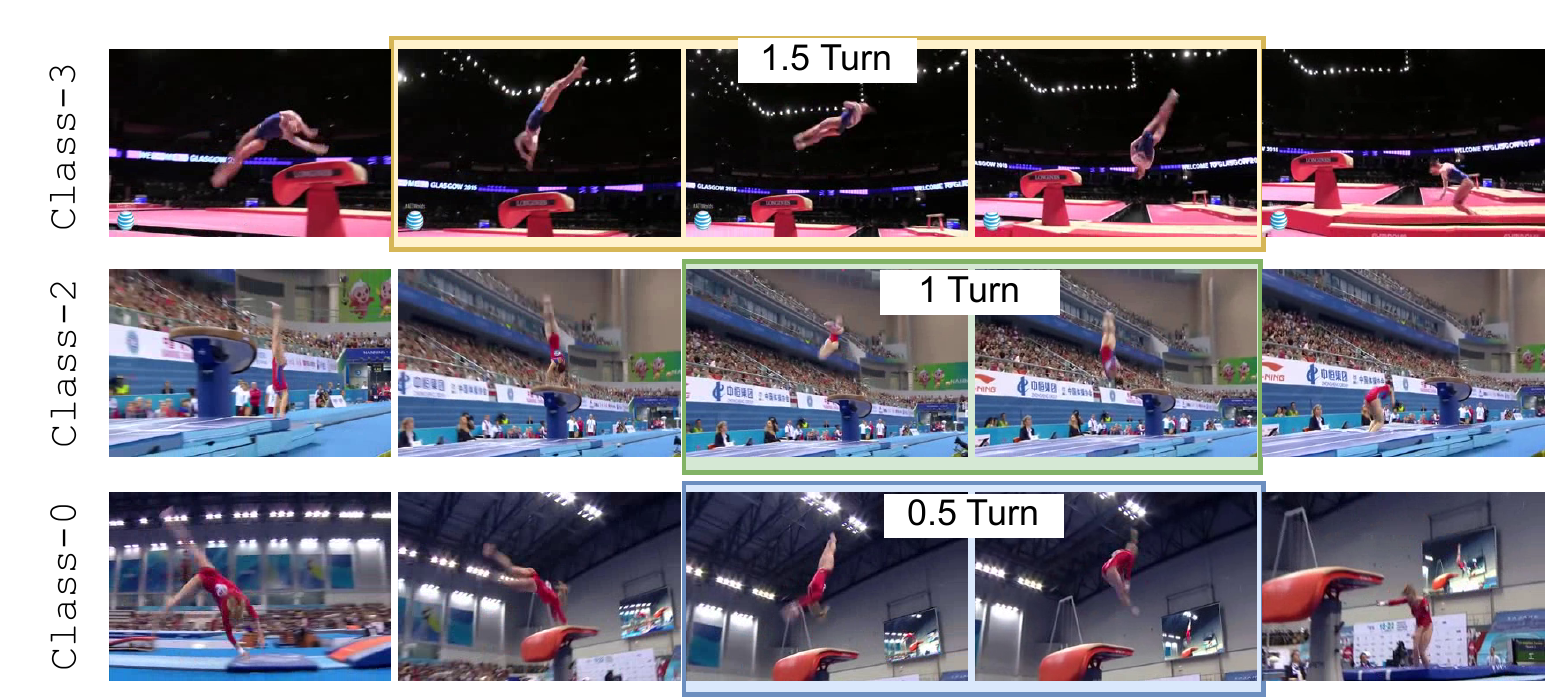}
    \caption{Samples from the FineGym Dataset. FineGym offers a range of challenging, fine-grained action classes derived from gymnastic events. This figure showcases three action classes from the FineGym288 split. Here, each action class differs in the phase where different numbers of turns are executed.}
    \label{fig:finegym_example}
\end{figure}

\noindent\textbf{FineGym}~\cite{shao2020finegym} provides challenging fine-grained action classes of various gymnastic events. Some samples from this dataset are shown in Fig.~\ref{fig:finegym_example}. Apart from FineGym99 and FineGym288 mentioned in the main paper, we also present results within each of the event subsets, as used in recent work~\cite{severe}.

\noindent\textbf{Vault (VT)}~\cite{shao2020finegym} contains 6 action classes from the Vault event. Its training/test split contains 1k/0.5k videos.

\noindent\textbf{Floor (FX)}~\cite{shao2020finegym} includes 35 action classes from the `Floor Exercise' event. Its training/test split contains 5.3k/2.2k videos.

\noindent\textbf{UB-S1}~\cite{shao2020finegym} comprises 15 action classes covering videos of different types of circles around the bars. Its training/test split contains 3.5k/1.5k videos.

\noindent\textbf{FX-S1}~\cite{shao2020finegym} is a subset of the Floor Exercise (FX) set, covering 11 actions related to leaps, jumps, and hops. Its training/test split contains 1.9k/0.7k videos.

\noindent\textbf{FineDiving}~\cite{xu2022finediving} includes approximately 3k videos covering 52 action classes from Diving sequences. This dataset focuses on the problem of action quality assessment, providing annotations for steps and scores. However, we utilize only the `action' annotations in our work.

\noindent\textbf{Kinetics400}~\cite{kinetics} contains more general human actions collected from YouTube. It covers 400 action classes, with a training/validation split of 240k/20k videos.

\noindent\textbf{Something-Something V2}~\cite{goyal2017something} focuses on actions related to hand-object interactions. We utilize a split from prior work~\cite{semi_tcl, semi_actorcutmix}, which covers 82k training and 12k test videos.

\section{Implementation Details}
\label{suppsec:implementation}

\noindent\textbf{Network Architecture}: Alignability Encoder ($f_A$) is a Video Transformer Network (VTN)~\cite{vtn} architecture following prior work~\cite{carl,vsp}. For non-linear project head $g(\cdot)$ we employ a multilayer perception (MLP) following~\cite{simclr}. For Action Encoder ($f_E$) we utilize the R2plus1D-18~\cite{r2plus1d} model by default, which is initialized with SSL pretraining~\cite{tclr} on the given dataset. For the score mapping function $f_S$ we utilize a 2-layer MLP.

\noindent\textbf{Training}: SSL pretraining of $f_A$ takes place for 100 epochs. Alignability-verification based metric learning of $f_A$ and training of $f_E$ takes 100 epochs. 
In the self-training steps, the proposed collaborative PL generation each takes place at every 5th epoch of labeled training, this process runs for 10 training iterations. 

\noindent\textbf{Inference} For inference, we only consider the video encoder $f_E$, following a commonly used protocol~\cite{r2plus1d}. We first obtain clip-level predictions from 10 uniformly sampled clips across the video duration and 3 spatial crops, then average these predictions to derive a video-level prediction. 
\subsection{Hyperparameters}
\subsubsection{SSL pretraining of $f_A$}
For the Gaussian Infused Temporal Distinctiveness Loss (\( \mathcal{L}_{GITDL} \)) (Eq.~\ref{eq:gitdl}), we set the temperature parameter (\( \tau \)) to 0.1. Additionally, for the Gaussian prior, we use a peak value (\( \kappa \)) of 0.99 and a standard deviation (\( \sigma \)) of 0.2.

\subsubsection{Alignability-based Metric Learning} 
After SSL pretraining of \( f_A \), we freeze the image encoder and continue training only the temporal encoder of the VTN architecture. For the computation of softDTW, we set the smoothness parameter (\( \gamma \)) to 0.001. In the case of the Alignability-based Triplet Loss (\( \mathcal{L}_{AT} \)), we use a default margin (\( m \)) of 0.1. Our batch size is set to 96, and we employ a subsampler in the dataloader to ensure that there are at least two instances from each sampled action class.

\subsubsection{Collaborative Pseudolabeling process}
To construct the embedding set \( \mathbb{A} \) from the labeled dataset, we randomly select \( \rho \) = min(15, samples in the class) samples from each class. For the non-parametric classifier (as detailed in \supp{Eq.~8 of the main paper}), we set the temperature parameter \( \tau \) to 0.1. The confidence threshold \( \theta \) is established at 0.6.

For our collaborative pseudo-labeling process, only a single forward pass is sufficient for each video in both $\mathbb{D}_l$ and $\mathbb{D}_u$ to extract their respective features. Subsequently, the classwise alignability score is computed in parallel on these extracted features, significantly enhancing the speed of the pseudo-labeling process and not bottlenecking the speed of the overall PL process.

\subsection{Optimization and Training Schedule}

To update the parameters of the network, we employ the Adam optimizer~\cite{adam}, using its default parameters, \(\beta_1 = 0.9\) and \(\beta_2 = 0.999\). For the learning rate scheduler, we apply a base learning rate of \(10^{-4}\), accompanied by a linear warmup over the first 5 epochs, followed by a cosine decay learning rate scheduler.

\section{Additional Ablations}
\label{suppsec:ablation}
For additional ablation, we follow the same default setup of the ablation of the main paper \ie reporting results on the action recognition task of various fractions of labeled set of Diving48 dataset with R2plus1D model.

\subsection{Ablation with Triplet Mining Strategies}
For our mini-batch sampling, we ensure that each class sampled has at least two instances. We then calculate the alignment cost (as per \supp{Eq. 1 in the main paper}) between each pair of samples within the mini-batch. Samples from the same class serve as positives, while pairs from different classes are considered negatives. While all positive pairs are included in our analysis, we explore various strategies for mining negative pairs in Table~\ref{abl:triplet}.
\begin{table}[h]
\centering
\caption{Ablation with Triplet loss}

\begingroup
\resizebox{0.5\linewidth}{!}{
\setlength{\tabcolsep}{4pt}
\begin{tabular}{lcc} 
    \arrayrulecolor{black}\hline
    
    \hline
    
    \hline\\[-3mm]
    \textbf{Triplet Mining} & \textbf{10\%} & \textbf{20\%}   \\ 
    \hline
    All Negatives                 & 36.16             & 58.65           \\
    Hard Negatives                & \textbf{37.64}    &  \textbf{60.40}  \\
    Hardest Negative only       & 37.20             & \textbf{60.40}           \\
    \bottomrule
\end{tabular}
}
\endgroup

\label{abl:triplet}
\end{table}

In the first row, where all negative pairs are considered, we observe less effective learning. This is due to easy negatives (where \( D^n - D^p < m \)) that fail to effectively contribute significantly to the learning process. On the other hand, mining hard negatives—specifically, considering only those negative pairs where \( D^n - D^p > m \)—and selecting the hardest negative from the mini-batch, shows improved performance. However, the `hardest-negative' strategy performs slightly worse than the `hard negatives' in the 10\% data scenario, likely due to the reduced number of available triplets.

\subsection{Empirical evidence: suitability of Alignment based distance}
We conduct experiments utilizing various distance functions $D$ in \supp{Eq. 2 of the main paper} to train $f_A$ using our proposed metric learning approach. Given that our metric learning is centered around a verification task (determining whether a video pair belongs to the same class), we also report the validation average precision in Table~\ref{table:distances_alignability}. The findings reveal that the alignment cost(softDTW) markedly outperforms other distance measures across diverse tasks. Moreover, for fine-grained action categories, a distance function based on alignment is far more effective than the standard cosine distance, underscoring our motivation \supp{Fig. 1(c) of the main paper}.

\begin{table}[h]
\caption{Ablation of different distance in metric learning}

\centering
\begingroup
\resizebox{0.48\linewidth}{!}{
\setlength{\tabcolsep}{2pt}
        \begin{tabular}{lccc} 
                    \arrayrulecolor{black}\hline
                    
                    \hline
                    
                    \hline\\[-3mm]
                    \textbf{Distance Type}              & \textbf{AP} & \textbf{10\%}  & \textbf{20\%}   \\ 
                    \hline
                    cosine- mean           & 0.57        & 33.41         & 53.60           \\
                    cosine- full seq.      & 0.48        & 32.15         & 52.54          \\
                    cosine- 4 seg          & 0.64        & 34.90          & 54.51          \\
                OTAM~\cite{otam}                   & 0.68        & 35.06         & 56.33          \\
                    softDTW~\cite{softdtw}                & \textbf{0.72}        & \textbf{37.64}         & \textbf{60.40}    \\
                    \bottomrule
                    \end{tabular}
        }
        \endgroup
        \label{table:distances_alignability}

\end{table}

\noindent\textbf{SSL pretraining of Alignability-encoder $f_A$:}

\begin{table}[h]
    \caption{Ablation: SSL pretraining of \( f_A \)}

\arrayrulecolor{black}
\centering
\arrayrulecolor[rgb]{0.753,0.753,0.753}
\begingroup
\setlength{\tabcolsep}{1pt}
\begin{tabular}{c!{\color{black}\vrule}c|c!{\color{black}\vrule}cc} 
\arrayrulecolor{black}\hline

\hline

\hline\\[-3mm]
\multirow{2}{*}{\textbf{SSL}} & \multicolumn{2}{c|}{\textbf{PennAction}} & \multicolumn{2}{c}{\textbf{Diving48}} \\
\arrayrulecolor[rgb]{0.753,0.753,0.753}\cline{2-3}
\multirow{2}{*}{\textbf{Objective}} & \textbf{Phase} & \textbf{Event} & \multirow{2}{*}{\textbf{10\%}} & \multirow{2}{*}{\textbf{20\%}} \\
& \textbf{Classi.} & \textbf{Progress} & & \\
\arrayrulecolor{black}\hline
w/o gaussian & 0.88 & 0.87 & 35.42 & 58.81 \\
with gaussian & \textbf{0.93} & \textbf{0.91} & \textbf{37.64} & \textbf{60.40} \\
\arrayrulecolor{black}\bottomrule
\end{tabular}
\endgroup
\label{table:alignssl}
\end{table}

To assess the representation quality of SSL pretraining of $f_A$ (\supp{Supp. Sec. E}), we conduct additional evaluations on fine-grained video tasks of the PennAction dataset~\cite{pennaction}: Phase Classification and Event Progress, following the protocol in~\cite{tcc}. These tasks are action-class agnostic and require an understanding of the action phase.

Results from Table~\ref{table:alignssl} suggest that our proposed Gaussian prior-based frame-level temporal distinctiveness significantly improves the performance of phase-level tasks and the overall video-level semi-supervised learning performance on fine-grained actions. This improvement is attributed to the Gaussian prior, which enhances temporal coherence (smoothness) in the frame-wise video embedding.

\section{Additional Results}
\label{suppsec:results}
\subsection{Results with ImageNet Pretraining}
We additionally present results using the ViT-B backbone, pretrained on ImageNet~\cite{deng2009imagenet}, and apply it to both fine-grained (Diving48) and coarse-grained (Kinetics400) datasets. These results are presented in Table~\ref{table:imagenet}. Our method surpasses the performance of the previous approach~\cite{xing2023svformer}, which employs the same backbone and pretrained weights.

\begin{table}[h]
\centering
\caption{Results with backbone initialization from ImageNet (supervised)}
\arrayrulecolor{black}

\begingroup
\setlength{\tabcolsep}{4pt}
\begin{tabular}{lcccccc}
\hline

\hline

\hline\\[-3mm]
\textbf{Method} & \textbf{Backbone} & \multicolumn{2}{c}{\textbf{Diving48}} & \multicolumn{2}{c}{\textbf{Kinetics400}} \\
                &                   & \textbf{10\%} & \textbf{20\%} & \textbf{1\%} & \textbf{10\%} \\
\hline
SVFormer-B~\cite{xing2023svformer}      & ViT-B             & 49.7         & 71.1          & 49.1        & 69.4          \\
Ours            & ViT-B             & \textbf{54.2}         & \textbf{75.7 }         & \textbf{52.0 }       &\textbf{ 71.1 }         \\
\hline

\hline

\hline\\[-3mm]
\end{tabular}
\endgroup
\label{table:imagenet}
\end{table}

\begin{table}[t]
\centering
\renewcommand{\arraystretch}{1.05}
\caption{Comparison with prior work of fine-grained video understanding on Action Recognition task.}

\begingroup
\resizebox{\linewidth}{!}{
\setlength{\tabcolsep}{2pt}
\begin{tabular}{lccccc}
\hline

\hline

\hline\\[-3mm]
\textbf{Method}          & \textbf{\% labels} & \textbf{Model}       & \textbf{Init. Data} & \textbf{FG99} & \textbf{FG288}  \\ 
\hline
$D^{3}$TW~\venueTT{CVPR'19}~\cite{d3tw}     & 100\%     & R(2D+3D)-50 & Labeled ImageNet & 15.3      & 14.1        \\
SpeedNet~\venueTT{CVPR'20}~\cite{speedNet} & 100\%     & R(2D+3D)-50 & Labeled ImageNet & 16.9      & 15.6        \\
TCN~\venueTT{ICRA'18}~\cite{tcn}      & 100\%     & R(2D+3D)-50 & Labeled ImageNet & 20.0      & 17.1        \\
SaL~\venueTT{ECCV'16}~\cite{misra2016shuffle}      & 100\%     & R(2D+3D)-50 & Labeled ImageNet & 21.5      & 19.6        \\
TCC~\venueTT{CVPR'19}~\cite{tcc}      & 100\%     & R(2D+3D)-50 & Labeled ImageNet & 25.2      & 20.8        \\
GTA~\venueTT{CVPR'21}~\cite{gta}      & 100\%     & R(2D+3D)-50 & Labeled ImageNet & 27.8      & 24.2        \\
CARL~\venueTT{CVPR'22}~\cite{carl}     & 100\%     & VTN (R50)   & Unlabeled ImageNet & 41.8      & 35.2        \\
VSP~\venueTT{CVPR'23}~\cite{vsp}    & 100\%     & VTN (R50)   & Unlabeled ImageNet & 43.1      & 36.9        \\
VSP-P~\venueTT{CVPR'23}~\cite{vsp}    & 100\%     & VTN (R50)   & Unlabeled ImageNet & 44.6      & 38.2        \\
VSP-F~\venueTT{CVPR'23}~\cite{vsp}    & 100\%     & VTN (R50)   & Unlabeled ImageNet & 45.7      & 39.5        \\
Ours(\textit{FinePseudo})           & 5\%       & VTN (R50)   & Unlabeled ImageNet & 41.1      & 34.4        \\
\rowcolor[rgb]{0.784,0.902,0.976}Ours(\textit{FinePseudo})           & 10\%      & VTN (R50)   & Unlabeled ImageNet & \textbf{66.2}      & \textbf{56.5}        \\
\hline

\hline

\hline\\[-3mm]
\end{tabular}
}
\endgroup
\label{table:FGonly}
\end{table}

\subsection{Results on FineGym subsets}
Results on the Standard FineGym99/288 splits, which encompass all four types of gymnastic events—Vault, Floor Exercise, Balance Beam, and Uneven Bars—are presented. The action classes from these diverse events are semantically distinct from one another. In our analysis, we treat actions from each event separately, adding further complexity to the classification problem. The results are detailed in Table~\ref{table:FG_withinset}. Initially, we evaluate video self-supervised learning baselines: TCLR~\cite{tclr} and VideoMoCo~\cite{videomoco}. Subsequently, models initialized with the weights from ~\cite{tclr} are used to assess semi-supervised methods. Our method consistently outperforms previous methods by a significant margin across all splits. This indicates the superior ability of our semi-supervised approach to distinguish fine-grained, semantically similar actions within each event set.

\begin{table}
\centering
\arrayrulecolor{black}
\caption{Results on within set activities of FineGym dataset}
\vspace{-3mm}
\begingroup
\resizebox{\linewidth}{!}{
\setlength{\tabcolsep}{1.5pt}
\begin{tabular}{l|ccc|ccc|ccc|ccc} 
\arrayrulecolor{black}\hline

\hline

\hline\\[-3mm]
\multirow{2}{*}{\textbf{Method}} & \multicolumn{3}{c|}{\textbf{Vault (VT)}}      & \multicolumn{3}{c|}{\textbf{Floor (FX)}}      & \multicolumn{3}{c|}{\textbf{UB-S1}}           & \multicolumn{3}{c}{\textbf{FX-S1}}             \\ 
\cline{2-13}
                                 & \textbf{5\%}    & \textbf{10\%}   & \textbf{20\%}   & \textbf{5\%}    & \textbf{10\%}   & \textbf{20\%}   & \textbf{5\%}    & \textbf{10\%}   & \textbf{20\%}   & \textbf{5\%}    & \textbf{10\%}   & \textbf{20\%}    \\ 
\hline
TCLR~\venueTT{CVIU'22\cite{tclr}}                            & 34.2          & 39.7          & 41.6          & 24.0          & 25.4          & 57.6          & 22.5          & 41.7          & 60.6          & 17.9          & 21.6          & 34.8           \\
VidMoCo~\venueTT{CVPR'21\cite{videomoco}}                        & 32.0          & 38.9          & 40.7          & 22.3          & 23.6          & 55.1          & 19.8          & 40.3          & 59.2          & 14.6          & 18.9          & 32.5           \\
\hline
PL                               & 34.1          & 39.9          & 42.4          & 23.9          & 25.7          & 58.1          & 22.8          & 42.3          & 62.5          & 17.4          & 21.5          & 35.1           \\
TimeBal~\venueTT{CVPR'23\cite{timebalance}}                      & 35.7          & 40.4          & 43.1          & 24.6          & 26.3          & 59.7          & 28.6          & 43.1          & 63.2          & 19.2          & 22.3          & 35.5           \\
\rowcolor[rgb]{0.784,0.902,0.976}Ours(\textit{FinePseudo})                              & \textbf{40.8} & \textbf{44.0} & \textbf{47.6} & \textbf{29.2} & \textbf{30.0} & \textbf{63.6} & \textbf{32.4} & \textbf{46.5} & \textbf{67.4} & \textbf{23.5} & \textbf{27.7} & \textbf{39.2}  \\
\bottomrule
\end{tabular}
}
\endgroup

\label{table:FG_withinset}
\vspace{-6mm}
\end{table}

\subsection{Comparison with fine-grained video methods}
Additionally, we compare our results with previous methods that specialize in video fine-grained intra-video tasks, as shown in Table~\ref{table:FGonly}. Without the need for extra data, our method surpasses these prior approaches by leveraging only 10\% of the labeled data.

\subsection{Results on Class-agnostic Fine-grained tasks}
While our primary focus is on semi-supervised action recognition, we also present the performance of our alignability encoder \( f_A \) on class-agnostic fine-grained tasks such as Phase Classification, Kendall's Tau, and Event Progress, as proposed by ~\cite{tcc}. We evaluate \( f_A \) directly following SSL pretraining, without the use of any labeled data. The results, detailed in Table~\ref{table:pennaction}, demonstrate that our method performs favorably compared to those specialized in these tasks. It also shows the effectiveness of our GITDL-based SSL pretraining in capturing tasks that are based on intra-video dynamics, such as action-phases.

\begin{table}[h]
\centering
\begingroup
\caption{Results on fine-grained tasks of PennAction dataset~\cite{pennaction}.}
\label{table:pennaction}

\setlength{\tabcolsep}{4pt}
\begin{tabular}{lcccc}
\hline

\hline

\hline\\[-3mm]
\textbf{Method} & \textbf{Label} & \textbf{Phase} & \textbf{Kendall's} & \textbf{Event} \\ 
                 &  \textbf{Used}                   & \textbf{Classi.} & \textbf{Tau} & \textbf{Progress} \\
\hline
TCC~\venueTT{CVPR'19}~\cite{tcc} & Action & 0.744 & 0.641 & 0.591 \\
GTA~\venueTT{CVPR'21}~\cite{gta} & Action & -     & 0.748 & -     \\
LAV~\venueTT{CVPR'21}~\cite{lav} & Action & 0.786 & 0.684 & 0.625 \\
\hline
SaL~\venueTT{ECCV'16}~\cite{misra2016shuffle} & None  & 0.682 & 0.474 & 0.390 \\
TCN~\venueTT{ICRA'18}~\cite{tcn} & None  & 0.681 & 0.542 & 0.383 \\
CARL~\venueTT{CVPR'22}~\cite{carl} & None  & 0.931 & 0.985 & 0.918 \\
VSP~\venueTT{CVPR'23}~\cite{vsp} & None  & 0.931 & 0.986 & \textbf{0.923} \\
\rowcolor[rgb]{0.784,0.902,0.976}Ours ($f_A$)         & None  & \textbf{0.932} & \textbf{0.992} & 0.911 \\
\hline

\hline

\hline\\[-3mm]
\end{tabular}
\endgroup
\end{table}

\subsection{Complementary behavior- VideoSSL methods}
To substantiate the claims, we analyze two distinct types of video SSL methods: (1) TCLR, which focuses on learning video-level representations for high-level semantic tasks such as action recognition, and (2) CARL, oriented towards learning frame-level video representations for low-level intra-video tasks like phase classification.

In our analysis, we utilize publicly available Kinetics400 pre-trained weights for both TCLR and CARL. We then evaluate their performance on intra-video tasks using the PennAction dataset~\cite{pennaction} and on the video-level action recognition task with the Diving48 dataset~\cite{diving}, as detailed in Table~\ref{table:abl_compl_videossl}. This comparison reveals distinct behavioral patterns of the two video SSL methods across these tasks.

\begin{table}[h]
\centering
\caption{Complementary behavior of VideoSSL methods.}

\begingroup
\setlength{\tabcolsep}{4pt}
\begin{tabular}{l|cc|cc} 
\hline

\hline

\hline\\[-3mm]
\multirow{2}{*}{\textbf{Method}} & \multicolumn{2}{c|}{\textbf{PennAction}}      & \multicolumn{2}{c}{\textbf{Diving48}}  \\
                                 & \textbf{Phase Class.} & \textbf{Kendall's Tau} & \textbf{10\%} & \textbf{20\%}          \\ 
\hline
CARL~\cite{carl}                             & 0.931                & 0.985                  & 26.8          & 47.1                   \\
TCLR~\cite{tclr}                             & 0.799                & 0.821                  & 33.1          & 53.7                   \\
\bottomrule
\end{tabular}
\endgroup
\label{table:abl_compl_videossl}
\end{table}

\section{Method}
\label{suppsec:method}
\subsection{SSL pretraining of Alignability encoder}
\label{sec:alignmentssl}
Given the limited scale of labeled data ($\mathbb{D}_{l}$), our primary objective is to effectively utilize the extensive scale of unlabeled data ($\mathbb{D}_{u}$) to facilitate the learning of frame-wise video representations in \(f_A\), which can useful to identify the action phase. 

Recent advancements in clip-level video self-supervised methods, have shown promising results in learning powerful representations within a single video instance by employing a temporal-distinctiveness objective~\cite{tclr, timebalance}. In the standard \textit{clip-level} temporal-distinctiveness formulation, within a video instance, temporally-aligned clips are treated as positive, while temporally-misaligned clips are considered negative. However, this approach treats each misaligned timestamp equally negative, regardless of their temporal distance from the anchor clip. 
In the context of \textit{frame-level} video representations, treating negative equally loses the frame-wise temporal coherence (smoothness). As established by prior work~\cite{carl, lav,vsp,gta,tcc}, it is crucial for capturing intra-video dynamics, such as action phases.
To address this and achieve temporal coherence in learning frame-wise temporal-distinctiveness, we introduce a gaussian kernel to the negative timestamps. This modification ensures that the weight of a negative instance increases smoothly (due to gaussian) and proportionally with its timestamp difference from the anchor.
\begin{figure}[h]
    \centering
    \includegraphics[width=0.7\linewidth]{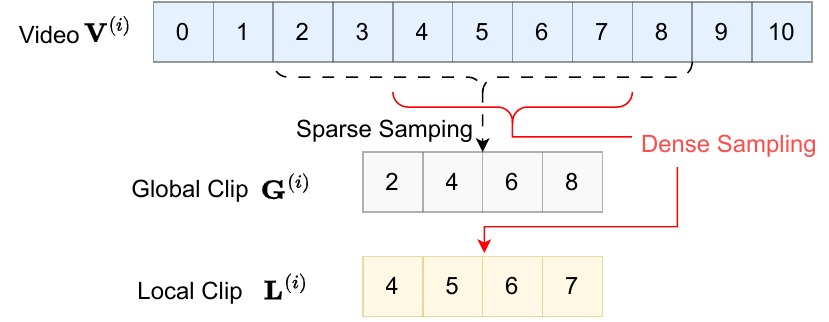}
    \caption{Clip Sampling in the Proposed GITDL Framework. From a full video \(\mathbf{V}^{(i)}\), we sample two types of clips: a global clip \(\mathbf{G}^{(i)}\), which is sparsely sampled (skip rate = 2), and a local clip \(\mathbf{L}^{(i)}\), which is densely sampled (skip rate = 1) within the temporal range of \(\mathbf{G}^{(i)}\).}
    \label{fig:gitdl_aid}
\end{figure}

Consider a video instance $i$ from which we sample a global clip $G$ and a local clip $L$, with $L$ being a subset of $G$. Both clips are sampled to have exactly $T$ frames - $G$ through sparse sampling and $L$ through dense sampling (Visual Aid in Fig.~\ref{fig:gitdl_aid}). These clips are then fed into the alignability encoder \(f_A\) and a non-linear projection layer \(g(\cdot)\), resulting in their frame-wise video representations $\{\bar{\mathbf{g}}_t^i\}_{t=1}^T$ and $\{\bar{\mathbf{l}}_t^i\}_{t=1}^T$. 
Next, we subsample these representations to retain only the frame-ids present in both clips. This results in temporally corresponding representations with $\mathcal{T}$ frames $\{{\mathbf{g}}_t^i\}_{t=1}^\mathcal{T}$ and $\{{\mathbf{l}}_t^i\}_{t=1}^\mathcal{T}$. Our novel objective, Gaussian Infused Temporal Distinctiveness Learning(GITDL), is formulated as follows:
\begin{equation}\label{eq:gitdl}
    \mathcal{L}_{GITDL}^{(i)}\! = \!-\!\sum_{t_1=1}^\mathcal{T}\!\log \frac{\mathrm{h}(\mathbf{l}_{t_1}^{(i)}, \mathbf{g}_{t_1}^{(i)})}{\sum\nolimits_{\substack{t_2=1 \\ t_2 \neq t_1}}^\mathcal{T} (1- \kappa e^{-\frac{(t_1-t_2)^2}{2\sigma^2}}) \mathrm{h}(\mathbf{l}_{t_1}^{(i)}, \mathbf{g}_{t_2}^{(i)})}
\end{equation}

\noindent Where $\mathrm{h}(\mathbf{u_{1}}, \mathbf{u_{2}})=\exp \left(\frac{\mathbf{u_{1}}^{T}\mathbf{u_{2}}}{\|\mathbf{u_{1}}\| \|\mathbf{u_{2}}\| \tau} \right)$ denotes the function for computing the similarity between the vectors $\mathbf{u_{1}}$ and $\mathbf{u_{2}}$, and includes a temperature parameter $\tau$. $\kappa$ and $\sigma$ denote the peak value and variance of the gaussian kernel.

We also present an ablation study on phase classification and overall action recognition in~\supp{Supp. Sec. C}.

\subsection{Open-World Semi-Supervised Learning}

\noindent \textbf{Standard Semi-Supervised Framework:}
In the standard semi-supervised action recognition framework, the dataset consists of two sets:

\begin{itemize}
    \item \textbf{Labeled Set} (\(\mathbb{D}_{l}\)): Includes video instances \(\mathbf{v}^{(i)}\) and their corresponding action labels \(\mathbf{y}^{(i)}\), from a set of predefined classes \(C\). 
    
    Formally, \(\mathbb{D}_{l} = \{(\mathbf{v}^{(i)}, \mathbf{y}^{(i)})\}_{i=1}^{N_{l}}\).
    \item \textbf{Unlabeled Set} (\(\mathbb{D}_{u}\)): Contains unlabeled video instances that are assumed to belong to the same set of classes \(C\). 
    
    It is defined as \(\mathbb{D}_{u} = \{\mathbf{v}^{(i)}\}_{i=1}^{N_{u}}\).
\end{itemize}

\noindent \textbf{Open-World Extension:}
In the open-world semi-supervised learning framework, we introduce the presence of novel action classes within the unlabeled data:

\begin{itemize}
    \item \textbf{Labeled Set}: Remains unchanged, with instances from the known classes \(C\).
    \item \textbf{Unlabeled Set} (\(\mathbb{D}_{u}'\)): Now includes instances from both the known classes \(C\) and additional novel classes \(C_{novel}\). Thus, samples in \(\mathbb{D}_{u}'\) may belong to either \(C\) or \(C_{novel}\). Represented as \(\mathbb{D}_{u}' = \{\mathbf{v}^{(i)}\}_{i=1}^{N_{u}'}\).
\end{itemize}

The objective is to improve action recognition for classes in \(C\) using both \(\mathbb{D}_{l}\) and \(\mathbb{D}_{u}'\), while effectively handling the label noise from novel class instances \(C_{novel}\) in \(\mathbb{D}_{u}'\).

\noindent \textbf{Experimental Setup:}
For our experiments (\supp{Sec 4.4 of main paper}) with the Diving48 dataset, 40 classes are designated as known classes \(C\) and the remaining 8 as novel classes \(C_{novel}\). This setup tests the model's ability to not only accurately recognize actions from the known classes using the available data but also adapt to the presence of novel class instances.

\section{Detailed Comparison to Prior Work}
\label{suppsec:priorwork}

\subsection{Utilization of Alignment-Based Objective in Limited Labeled Setup}
To the best of our knowledge, the work most closely related to ours in terms of utilizing an alignment cost is ~\cite{otam}, which employs alignment cost directly to match queries with a support set in few-shot procedural video classification.

Our approach, however, differs from~\cite{otam} significantly in several key aspects:
\begin{enumerate}
    \item \textbf{Focus on Temporally Fine-Grained Actions:} We target temporally fine-grained actions where learning action phases is crucial, as opposed to procedural videos. Additionally, our semi-supervised framework leverages a substantial amount of unlabeled data, whereas ~\cite{otam} confines itself to a few-shot learning setup without using unlabeled data.
    \item \textbf{Application of Alignability Score:} Instead of directly using the alignment cost for classification, we introduce a learnable alignability score to address a binary classification problem, encouraging a focus on intra-video features. Our concept of `alignability' (determining if two clips are alignable) contrasts with the approach in ~\cite{otam}, which applies alignment cost for multi-class classification.
    \item \textbf{Temporal Context and Encoder Design:} ~\cite{otam} relies on a frame-level encoder and attempts frame-level alignment without temporal context. In contrast, our approach employs a frame-wise video encoder, pretrained with GITDL to grasp action phases before computing the alignment cost, thereby integrating temporal context into the model.
    \item \textbf{Variant of DTW in ~\cite{otam}:} The study in ~\cite{otam} introduces an interesting variant of Dynamic Time Warping (DTW) with relaxed boundary conditions to find the optimal path of alignment. We explore this variant in our ablation study (Table~\ref{table:distances_alignability}). While it proves effective for procedural videos, in our context of temporally fine-grained actions, we observe that it performs less effectively than the regular DTW.
\end{enumerate}

\subsection{SSL Pretraining - GITDL}
To utilize the unlabeled set \(\mathbb{D}_u\) for learning a frame-wise video encoder \( f_A \) that focuses on intra-video dynamics such as action phases, we introduce the Gaussian Infused Temporal Distinctiveness Loss (GITDL). The most closely related SSL pretraining methods to our GITDL are ~\cite{carl} and ~\cite{vsp}.
\begin{table}
\centering
\caption{Different SSL Objectives for Alignability encoder}

\begingroup
\setlength{\tabcolsep}{4pt}
\begin{tabular}{lcc} 
\hline

\hline

\hline\\[-3mm]
\textbf{SSL pretraining of $f_A$} & \textbf{10\%} & \textbf{20\%}  \\ 
\hline
CARL                           & 36.2          & 59.3           \\
GITDL                          & \textbf{37.6}          & \textbf{60.4}           \\
\bottomrule
\end{tabular}
\endgroup
\label{table:carl_abl}
\end{table}

Key differences include:
\begin{enumerate}
    \item \textbf{Temporal Distinctiveness in GITDL vs. Temporal Invariance in CARL:} Our GITDL aims to learn explicit `temporal distinctiveness', contrasting with the SSL objective of CARL, which promotes `temporal invariance'. Mathematically, our loss (Eq.~\ref{eq:gitdl}) considers only temporally-aligned frames as positives, whereas~\cite{carl} treats all frames as positives (Eq. 1 of ~\cite{carl}), thereby fostering temporal invariance. Moreover, we apply a Gaussian prior to the negatives of the anchor, while CARL treats all negatives uniformly.
    
    \textbf{Video as a Process in VSP:} VSP(\cite{vsp}) views videos as a process and learns through a Brownian bridge with a triplet loss, which differs from our GITDL.
    \item \textbf{Global and Local Clip Views:} Our approach incorporates both global and local views of a clip, providing more temporal context compared to the fixed-length clips in ~\cite{carl} and ~\cite{vsp}. This global perspective better suits the subsequent learning stages, particularly the video-level alignability-verification objective using labeled data.
\end{enumerate}

We have integrated the publicly available code of CARL (~\cite{carl}) into our framework for comparative analysis, shown in Table~\ref{table:carl_abl}. Although CARL yields impressive results on class-agnostic intra-video tasks (Table~\ref{table:pennaction}), it is slightly less effective in video-level semi-supervised tasks. Our conjecture is that this is due to the absence of global temporal context in ~\cite{carl} pretraining.

\subsection{Training Cost Comparison with Prior Work}

Our method only utilizes the single RGB modality, in contrast to methods like ~\cite{semi_mvpl, semi_tgfixmatch}, which employ additional modalities such as optical flow or temporal gradients. These extra modalities lead to a significant increase in training time due to two main factors: (1) the extended preprocessing time required to compute flow or temporal gradients, and (2) increased I/O overhead for loading both RGB and flow/gradient data. For example, computing optical flow for the Kinetics400 dataset can span several days and requires 3-5 terabytes of additional storage space. Conversely, our method efficiently operates on RGB-only videos, avoiding these extensive computational demands.

In terms of memory requirements, our framework is notably more efficient. Each branch of our model ($f_E$ and $f_A$) is trained independently, thereby reducing the overall memory consumption. This is in stark contrast to training frameworks like ~\cite{semi_cmpl, semi_videossl}, which necessitate running both teacher and student branches in training mode simultaneously, significantly increasing the memory footprint. Additionally, our approach does not require high-capacity teacher models. For instance, ~\cite{semi_cmpl} employs a 3D-ResNet50-4x width as a teacher, whereas ~\cite{timebalance, semi_cmpl} use two 3D-ResNet50 teachers. In comparison, our model efficiently utilizes only one teacher, further enhancing our method's resource efficiency.

% \bibliographystyle{splncs04}
% \bibliography{main}
\end{document}